\documentclass[letterpaper, 10 pt, conference]{ieeeconf}  

\IEEEoverridecommandlockouts                              
\let\labelindent\relax
\usepackage{enumitem}
\usepackage[utf8]{inputenc}
\usepackage[T1]{fontenc}
\usepackage{xcolor}
\usepackage{graphicx}
\usepackage{tabularx}
\usepackage{booktabs}
\usepackage{multirow}
\usepackage{colortbl}
\usepackage{dsfont}
\usepackage{algorithm}
\usepackage{algpseudocode} 

\usepackage{amsmath,amssymb,amsfonts}
\usepackage{graphicx}
\usepackage{subcaption} 

\newcommand*{\bx}{\mathbf{x}}
\newcommand*{\bv}{\mathbf{v}}
\newcommand*{\bb}{\mathbf{b}}

\title{\LARGE \bf
Efficient first-order predictor-corrector multiple objective optimization for fair misinformation detection
}

\author{Eric Enouen$^{1*}$ and Katja Mathesius$^{2*}$ and Sean Wang$^{3*}$ and Arielle Carr$^{4}$ and Sihong Xie$^{4}$
\thanks{$^{1}$The Ohio State University}
\thanks{$^{2}$Drake University}
\thanks{$^{3}$Cornell University}
\thanks{$^{4}$Computer Science and Engineering Department, Lehigh University}%
\thanks{$^{*}$ Denotes equal contribution.}
}

\begin{document}

\maketitle
\thispagestyle{empty}
\pagestyle{empty}

\begin{abstract}
	Multiple-objective optimization (MOO) aims to simultaneously optimize multiple conflicting objectives and has found important applications in machine learning,
	such as minimizing classification loss and discrepancy in treating different populations for fairness.
	At an optimality, further optimizing one objective will necessarily harm at least another objective,
	and decision-makers need to comprehensively explore multiple optima (called Pareto front) to pin-point one final solution.
	We address the efficiency of finding the Pareto front.
	First, finding the front from scratch using stochastic multi-gradient descent (SMGD)
		is expensive with large neural networks and datasets.
		We propose to explore the Pareto front as a manifold
		from a few initial optima, based on a predictor-corrector method.
	Second, for each exploration step, the predictor solves a large-scale linear system that scales quadratically in the number of model parameters,
		and requires one backpropagation to evaluate a second-order Hessian-vector product per iteration of the solver.
		We propose a Gauss-Newton approximation that only scales linearly,
		and that requires only first-order inner-product per iteration. This also allows for a choice between the MINRES and conjugate gradient methods when approximately solving the linear system. 
	 The innovations make predictor-corrector possible for large networks.
	Experiments on multi-objective (fairness and accuracy) misinformation detection task 
		show that 1) the predictor-corrector method can find Pareto fronts better than or similar to SMGD with less time;
		and 2) the proposed first-order method does not harm the quality of the Pareto front identified 
		by the second-order method, while further reduce running time. 
\end{abstract}

\section{Introduction}
Multi-objective optimization aims to find optimal solutions for multiple objective functions and has been an important tool for data mining and machine learning.
	For example, in multi-task learning, each learning task has an objective function to be optimized and tasks can be optimized jointly;
		in a recommendation system, content relevance and personalization are two important goals for the system to achieve simultaneously;
		in financial investment, uncertainty and loss are two objectives for an investor to minimize at the same time.
Usually the multiple objectives are conflicting and it is impossible to find a solution that is optimal for all individual objective functions.
Rather, trade-offs among the objectives are necessary and 
optimality in MOO can be characterized by the Pareto optimality:
an Pareto optimum is a solution where 
improving any one objective function necessarily harm at least another objective function, and jointly improving all objectives is impossible.

Previous work~\cite{Fliege2018} aims to find a single Pareto optimum without controlling the trade-off among the objectives.
This is undesirable 
since an objective function may not be sufficiently minimized, while the users cannot access and compare multiple trade-offs.
To address this, the authors of~\cite{Sener2018nips, Liu2019} proposed multi-gradient descent methods 
and can maintain a set of current best trade-offs and push them towards the Pareto front.
However, they still cannot control which trade-offs to reach during the optimization and the solutions can have similar objective function values.
In~\cite{Lin2019_nips, mahapatra20a}, further constraints are added to the gradient-based optimization
so that preferences over the trade-offs can be specified and the solutions are better spread across the front.

One common drawback of the above prior work is the computational efficiency.
One or a few optimal trade-offs is not sufficient, as practitioners need to adjust trade-offs in a fine-grainded manner to pinpoint the best trade-off. 
It is necessary to comprehensively traverse the the Pareto front.
The above optimization algorithms start from arbitrary initial points that can be quite far away from the fronts, and many steps must be taken to reach the fronts.
Furthermore, recovering the fronts can take many random initializations.
As the cost per iteration can be high with large dataset and neural networks, such methods are not feasible for recovering the Pareto fronts. 
The continuation algorithms, in particular the predictor-corrector methods, address this issue.
Intuitively, the optimal solutions are assumed to form a low dimensional manifold,
and near an optimal solution the manifold can locally be approximated by a plane described by a linear system.
Therefore, moving from one optimal solution to a neighboring optimal solution can be done
by first solving the approximating linear system (the predictor) to find an exploration direction.
Since the plane may not fully approximate the non-linear manifold,
moving along the exploration direction can get off the manifold.
The error can be corrected by a few steps of multi-gradient descent back to the manifold of optimal solutions.
Since the Pareto front is a low-dimensional manifold in the objective value space,
the predictor-corrector method is expected to travel less to identify the Pareto front.

However, in the predictor step, we still face two challenges centering around solving a large scale linear system in the form of $H\bv = \bb$,
where $H$ is an $n\times n$  symmetric matrix and $\bv\in \mathbb{R}^n$ is a vector of exploration direction that moves the current Pareto optimum $\bx$
to $\bx+\bv$, and
$\bb=\sum_{j=1}^{m}\nabla \beta_j f_j(\bx)$ is a linear combination of the gradients of individual objective functions $f_j$, $j=1,\dots, m$.
First, the system $H\bv=\bb$ requires an iterative solver, such as the Krylov methods conjugate gradient (CG) or MINRES \cite{VanDerVorst2003,Saad03}. 
As the system needs to be solved per predictor iteration, the cost associated with the solver can greatly impact the overall running time of exploring a Pareto front. 
Due to the unknown properties of the system under the specific context, currently no study about the efficiency of the solvers is available.
Since we do not explicitly store
the Hessian, $H$, the use of Krylov methods is a natural choice as such solvers require only a linear operator (or matrix-vector product) to construct the Krylov subspace.
In the present study, we consider CG and MINRES since both can be applied to symmetric coefficient matrices.  CG,
however, requires that the matrix be  symmetric {\it positive definite} (SPD) and while $H$ is guaranteed to be symmetric, we cannot immediately assume it is SPD.  When applying the
Gauss-Newton Hessian approximation as in  \cite{Gar2020} and based on the Levenberg–Marquardt method described in \cite{Noc2006}, we achieve an SPD approximation of $H$, affording the choice between CG and MINRES. 

Second, as indicated in~\cite{Martin2018,ma20aICML}, the matrix $H$ can be the linear combination of the Hessian matrices of the objective function.
Direct evaluation of the Hessian matrices can be costly or even infeasible for deep learning models.
One can apply the Pearlmutter trick and use a forward and backward propagation implemented by auto-differentiation to evaluate the left-hand-side $H\bv$.
This is still quite costly for large networks and datasets, since one forward-backward propagation is needed per iteration of the solver,
which is invoked to solve a linear system per iteration of the predictor to find the next Pareto optimum.
As a result, exploring the entire Pareto front can be time-consuming.

We propose a different approach to approximate the Pareto front, starting from an optimal solution and based on the predictor-corrector method.
Intuitively, the optimal solutions are assumed to form a low dimensional manifold,
and near an optimal solution the manifold can locally be approximated by a plane described by a linear system.
Therefore, moving from one optimal solution to a close-by neighboring optimal solution can be done by first solving the approximating linear system (the predictor) to find a direction.
Since the plane may not fully approximate the manifold if the move is far away, the approximation error can be corrected by a few steps of multi-gradient descent back to the manifold of optimal solutions (the Pareto front). 

The technical report is organized as follows.
In Section~\ref{sec:prelim}, we review the background of GNN-based detector and multi-objective optimization (MOO).
In Section~\ref{sec:method}, we describe the predictor-corrector algorithm.
In Section~\ref{sec:experiments}, we validate our claims through empirical experiments on three datasets for fake review detection.

\section{Preliminaries}
\label{sec:prelim}

In this section we review the basics of multi-objective optimization, predictor-corrector methods, and iterative methods for solving linear systems.
The notation is in Table~\ref{tab: notations}.
\begin{table}[t]
\small
\caption{Notations and definitions.}
\begin{tabular}{m{1.5cm}<{}m{6cm}<{}}  
\toprule
\multicolumn{1}{m{1.5cm}}{\textbf{Notations}} & \multicolumn{1}{m{6cm}}{\textbf{Definitions}}\\
\midrule
\midrule
$f_i$ & The $i$-th objective function to minimize\\
$\mathbf{f}$ & The vector function of the $m$ objectives\\
$\bx$ & Vector of parameters of $\mathbf{f}$\\
$J$ & The Jacobian of $\mathbf{f}$ w.r.t. $\bx$\\
$H_i$ & The Hessian matrix of $f_i$ w.r.t. $\bx$\\
$H^i$ & The $i^{th}$ power of $H$\\
$\mathbf{v}^{(i)}$ & The approximate solution at the $i^{th}$ iteration\\
$J^T$ & The transpose of a matrix, $J$\\
\bottomrule
\end{tabular}
\label{tab: notations}
\end{table}

\subsection{MOO and Multi-gradient descent}
We consider an MOO problem that has $m$ objective functions $f_i(\bx)$, $i=1,\dots, m$,
where $\bx\in \mathbb{R}^{n}$ is the parameter of the functions.
For example, in fair machine learning, the goal is to optimize a predictive model's parameter $\bx$ so as to minimize classification loss (measured by $f_1$) while reducing the discrepancy (measured by $f_2$) between the treatment of different populations. 
We let $\textbf{f}=[f_1,\dots, f_m]^\top:\mathbb{R}^n\to \mathbb{R}^m$ be the vector of the $m$ objective functions.
One would like to find an optimal $\bx^\ast$ that minimizes all the $m$ objectives simultaneously.
Since the objectives can be conflicting and no single solution $\bx$ can attain all the minima of individual objectives,
one has to resort to some trade-offs among the objectives.
To characterize the optimality with multiple conflicting objectives,
a Pareto optimum is a solution where simultaneously reducing all $m$ objectives is impossible, and reducing one objective will necessarily increase at least another objective.
We say that the solution $\bx$ dominates another solution $\bx^\prime$ if
$f_i(\bx)\leq f_i(\bx^\prime)$ for all $i=1,\dots, m$ and at least one strict inequality holds.
A Pareto optimum is optimal in the sense that it is not dominated by any other solutions.

There can be multiple Pareto optima and the image of set of Pareto optima under the mapping $\mathbf{f}$ in the space $\mathbb{R}^m$ is called the Pareto front.
The goal of many MOO problems is to generate a Pareto front for $\mathbf{f}$ so a user can select the optimal solution that has the desired trade-offs for the problem at hand.
For example, for fair machine learning, a user wants to find a predictive model with accuracy higher than a given threshold while minimizing unfairness.
Without searching for a Pareto front, the user may not be able to find solution with the desired levels of accuracy and fairness. 
To find a Pareto front, the multi-gradient descent algorithm~\cite{Fliege2000} starts from a random initial solution and for each iterate, calculates a descent direction that can jointly reduce all objective functions.
The iterations continue until a Pareto optimum is reached where such a descent direction is impossible.
For example, in~\cite{Fliege2000},
if the current solution $\bx$ is not on the Pareto front,
the following quadratic program optimizes the weights ($\lambda_i$, $i=1,\dots, m$) of the gradients of the objectives so that the linear combination of the gradients using the optimal weights is a descent direction for all objectives, 
\begin{align}
\max_\lambda & -\frac{1}{2}\left\Vert\sum_{i=1}^m \lambda_j(\nabla f_i(\bx))\right\Vert^2 \\
\textnormal{s.t. } & \sum_{i=1}^m \lambda_i = 1, \lambda_i >= 0, i=1,...,m,
\end{align}
where $\nabla f_i(\bx)$ is the gradient of the $i$-th objective function at the current solution $\bx$.

The above method rely on first-order information of the objectives and can be extended to find multiple Pareto solutions on the front~\cite{Liu2019}.
However, 
a Pareto solution is found based on a random initial solution, from which the method can take many steps to move to the Pareto front.
Furthermore, to recover a Pareto front, multiple search using the multi-gradient descent algorithm is required. 

\subsection{Predictor-corrector methods}
\label{sec:method}
\begin{figure}
    \centering
    \includegraphics[width=0.4\textwidth]{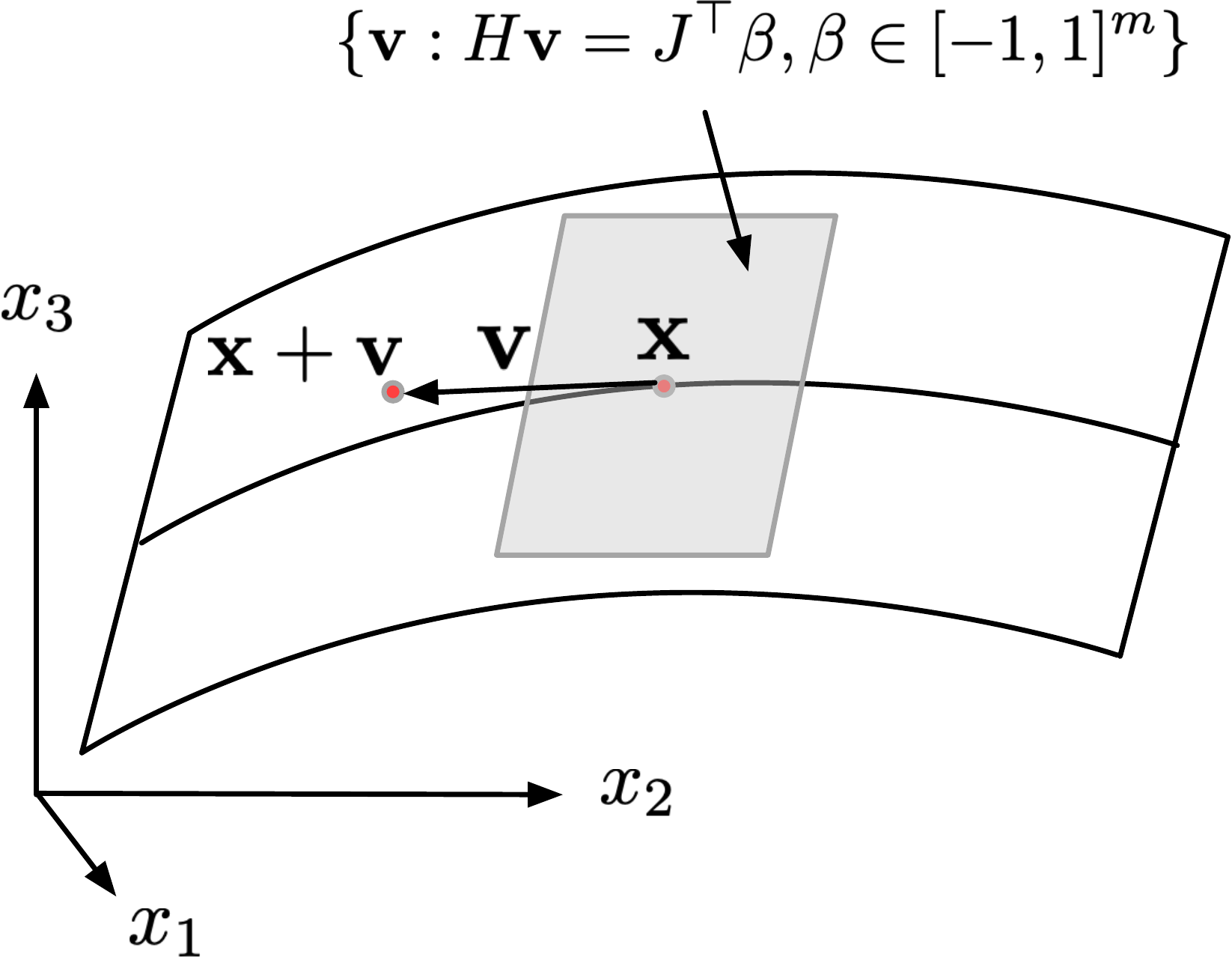}
    \caption{Pareto optimal solutions as a 2-dimensional manifold in $\mathbb{R}^3$,
    and the predictor can use the tangent plane to predict the direction $\mathbf{v}$ so that $\bx+\bv$ is close to the next Pareto optimum.}
    \label{fig:manifold}
\end{figure}
We introduce the predictor-corrector method that can find the Pareto front more efficiently. 
Predictor-corrector methods were introduced to MOO in~\cite{Martin2018,ma20aICML} as a way to explore a Pareto front, which is considered a low-dimensional manifold that can be approximated locally using a tangent plane.
A predictor-corrector method consists of two steps: a predictor step that approximately moves to a neighboring point of the current Pareto solution, and a corrector step that push the approximated neighboring point onto the Pareto front, where another round of prediction and correction can be conducted to generate the next Pareto solution. 


\noindent\textbf{Predictor}.
This step is derived in detail by \cite{ma20aICML}, so for brevity, we present only the main result below.
Locally, at the current Pareto solution $\bx$,
the hyperplane in $\mathbb{R}^n$ that touches $\bx$ and is tangent to the manifold can be described by the linear system 
\begin{equation}
\label{eq:predictor_direction}
    H(\bx)\textbf{v} = J^T \beta,
\end{equation}
where $H(\bx)=\sum_{i=1}^{m}\alpha_i H_i(\bx)$ is a linear combination of the Hessians of individual objectives, $\mathbf{v}$ represents the
exploration direction so that $\bx+\bv$ is on the tangent hyperplane,
and $\beta$ is a weighting vector chosen from $[-1, 1]^{m}$ to determine which objectives to increase/decrease by moving along $\bv$.
See Figure~\ref{fig:manifold}
for a demonstration with $\bx\in\mathbb{R}^3$.


\noindent\textbf{Corrector}.
The corrector step can employ any multi-objective optimization method, such as the above-mentioned multi-gradient descent method.
The idea is to make descent towards the Pareto front starting from $\bx+\bv$ where $\bv$ is the exploration direction generated by the predictor.
This step is necessary since the predictor approximate the manifold using a linear system and $\bx+\bv$ can be off the front.

\subsection{Krylov Subspace Methods}

A Krylov subspace method iteratively computes an approximate solution $\textbf{v}^{(j)}$ at iteration $j$ for the linear system
$H\textbf{v} = \textbf{b}$ by updating the initial solution $\textbf{v}^{(0)}$ as $\textbf{v}^{(j)} = \textbf{v}^{(0)}+\textbf{z}^{(j)}$, with
$\textbf{z}^{(j)}\in\mathcal{K}^{j}(H,\textbf{u})$.  Here, 
\begin{equation}\label{eq:krysubback}
    \mathcal{K}^{j}(H,\textbf{u}) = span{\{\textbf{u}, H\textbf{u},H^2\textbf{u},\dots,H^{j-1}\textbf{u}\} }
\end{equation}
is the Krylov space of dimension $j$ for $H\in\mathbb{C}^{n\times n}$ and $\textbf{u}\in\mathbb{C}^n$.\footnote{Notationally, we let $H$ be our coefficient matrix throughout, but note that this need not be a Hessian matrix.} The update
$\textbf{z}^{(j)}$ comes from a projection onto the Krylov space, and its computation, as well as the choice of $\textbf{u}$, is
specific to the method. We note that $\mathcal{K}^{j}$ is never explicitly computed as in (\ref{eq:krysubback}). For the methods
employed in this paper, MINRES implicitly builds an orthogonal basis for the Krylov space and iterates such that the residual
$\|\textbf{r}^{(j)}\| = \|\textbf{b} - H\textbf{v}^{(j)}\|$ is minimized.  CG does so such that the residual is orthogonal to
$\mathcal{K}^{j}$. Other approaches, such as biconjugate gradients and error minimizing methods also exist (see e.g., \cite{VanDerVorst2003}). More details on the Krylov methods specifically employed in this paper are given in Section
\ref{sec:minrescg}, and we refer the reader to \cite{VanDerVorst2003,Saad03} for further details on Krylov methods in general.

\section{Proposed method}
\subsection{Gauss-Newton Approximation}
\label{sec:gn}
The key bottleneck of the predict-ocrrector method is to solve the large-scale linear system in Eq.~\ref{eq:predictor_direction}.
There are challenges.
1) The matrix $H$ is large and dense to be computed and stored explicitly.
2) Inverting $H$ is very expensive as it takes $O(n^3)$ time complexity and iterative methods (Section~\ref{sec:minrescg}) are necessary.
3) Though an iterative method using the Pearlmutter trick can avoid the direct evaluation and inversion of $H$, it needs one backpropagation for every step of the iterative methods. With large datasets, especially those without the I.I.D. assumption to facilitate stochastic gradient estimation,
estimation of the matrix-vector product requires going through the datasets once for each backpropagation.
Lastly, the matrix $H$ is not necessarily positive semi-definite (PSD) to guarantee the success of certain iterative methods, such as conjugate gradient.

We propose the following techniques to avoid the above obstacles.
On the high level, our algorithm has the benefits of not evaluating the Hessian matrix and not even use backpropagation per iteration to evaluate a single Hessian-vector product.
The innovative techniques will rely on the
Gauss-Newton approximation~\cite{sagun2017empirical},
which uses a low-rank positive semi-definite matrix to approximate $H$ for optimizing a single scalar function $f(\bx)$.
Specifically for minimizing a classification loss $\ell=\ell(h(\bx))$ where $h$ is the machine learning model that output a real value for regression or a class probability distribution for classification,
and $\ell$ is a convex loss function, such as square error for linear regression or negative log-likelihood for classification, that takes $h(\bx)$ as input.
The Gauss-Newton approximation is based on the following
\begin{eqnarray}
    \ell^\prime & = & \ell^\prime(h(\bx)) \nabla_{\bx} h(\bx),\nonumber\\
    \ell'' & = & \nabla_{\bx} h(\bx) \ell''(h(\bx)) (\nabla_{\bx} h(\bx))^\top
            +\ell^\prime(h(\bx)) \nabla_{\bx}^2 h(\bx),\nonumber
\end{eqnarray}
where the second line follows from the chain rule of derivatives.
The Gauss-Newton approximation replaces the Hessian $H$ with  $\nabla_{\bx} h(\bx) \ell^{\prime^\prime}(h(\bx)) (\nabla_{\bx} h(\bx))^\top$.
This is reasonable when  $\ell^\prime(h(\bx))$ is near 0 or $h(\bx)$ is near linear around $\bx$~\cite{sagun2017empirical}.

Inspired by the above analysis, we will exploit the Gauss-Newton approximation to speed up solving Eq. (\ref{eq:predictor_direction}) within the predictor.
However, the objective functions that we adopt are loss functions that may not be convex with respect to the output of the model $h(\bx)$, and the term $\ell''(h(\bx))$ may not be positive to ensure that the Gauss-Newton approximation is positive definite.
For example, various fairness loss functions are non-convex~\cite{Wu2019}.
Another issue with the second-order term is when the loss is piece-wise linear, so that the term, and the Gauss-Newton approximation, becomes zero in some regions, leading to an invalid linear system in Eq. (\ref{eq:predictor_direction}).
We propose to use the following Gauss-Newton approximation for the $i$-th objective function $f_i$:
\begin{equation}
    \text{GN}_i = \nabla_{\bx} f(\bx)\nabla_{\bx}f(\bx)^\top = J_i^\top J_i,
\end{equation}
where the Jacobian matrix $J_i$ is the matrix of rows of partial derivative of $f_i(\bx)$ with respect to $\bx$.
Then the matrix $H(\bx)=\sum_{i=1}^n \alpha H_i(\bx)$ in Eq. (\ref{eq:predictor_direction}) is then replaced with
$\text{GN}(\bx)=\sum_{i=1}^n \alpha \text{GN}_i(\bx)$.
This approximation is guaranteed to be positive definite so long as not all gradients are zeros.
This property is beneficial to the iterative solvers, such as CG.

Regarding the computation complexity,
we are no long required to use backpropagation that go through the training data once to evaluate $H(\bx)\bv$ per iteration of the solvers.
Instead, we can go through training data just once to evaluate and then cache the Jacobian matrices, which scales only linearly $n$, the dimension of $\bx$, and in $m$, the number of objectives.
Per iteration, only $m$ vector inner product (evaluating $J_i \bv$) and $m$ scalar-vector product (evaluating $J_i^\top (J_i \bv)$) are needed.
The Jacobian matrices can be updated at the next iteration of the Predictor-Corrector method where one reaches the next Pareto optimum.

\subsection{CG and MINRES}\label{sec:minrescg}
In order to compute $\bv$ in (\ref{eq:predictor_direction}), we want to avoid explicitly taking the inverse of the Hessian (i.e., we avoid {\it directly} solving the system) since this is generally far too expensive for our purposes, particularly as the number of parameters
becomes very large.  Instead, we employ  computationally-less expensive iterative methods for approximately solving linear
systems, and in particular the Krylov methods MINRES and CG, which can be implemented in a matrix-free fashion.  In other words, we do not need to explicitly store the matrix in memory; rather, we simply need the matrix-vector product, or linear operator, defined. A major contribution of
previous work was introducing the use of the iterative solver, MINRES \cite{ma20aICML}. The residual is guaranteed to be monotonically decreasing in
MINRES, enabling early termination of the method. In the present study, we aim to expand the user's choice to include CG and we provide
a novel comparative analysis of these two iterative solvers when performing multiple-objective optimization.

CG and MINRES are iterative solvers for symmetric linear equations $H\bv=\bb$. CG
\cite{Hestenes1952} was developed as a more computationally efficient variant of the gradient descent method and specifically requires that $H$
also be positive definite. Both methods operate by finding the gradient at the current solution point and then moving in the in the conjugate
direction in the $H$-orthogonal direction. That is, at iteration $k$, given a point $\bv^{(k)}$ and a direction $\textbf{p}^{(k)}$, CG performs a
line search to find the value $\alpha$ to update the solution as $\bv^{(k+1)} = \bv^{(k)} + \alpha \textbf{p}^{(k)}$.
Then, a new direction that is conjugate to $\textbf{p}^{(k)}$ is computed such that at each iteration $\textbf{r}^{(k)}\perp \mathcal{K}^k$.
This method is guaranteed to converge in at most $n$ iterations when $H$ is an $n \times n $ matrix, but it is well-known that CG often
reaches an acceptable tolerance in far fewer iterations. For ease of reference, we provide the CG method following that in \cite{Saad03} in Algorithm \ref{alg:cg}.\footnote{Note that in the iterative method literature, $H$ is often used to denote an upper Hessenberg matrix, especially for methods like GMRES~\cite{Saad86}. For consistency, we use $H$ to denote a general coefficient for our application.}  

The MINRES method solves the system $H\bv=\bb$ by choosing the update to the approximate solution $\textbf{v}^{(k)}$ such that $\|\textbf{r}^{(k)}\| =\|\bb-H\textbf{v}^{(k)}\|$ is minimized. One of the key features of this method is due to the symmetry of
$H$: MINRES saves significant memory costs, requiring the storage of only the two previously computed basis vectors from the Krylov
space. This is performed via the Lanczos algorithm using what is referred to as a three-term recurrence; we omit the details here but refer
the reader to Section 6.6 in \cite{Saad03}. Since the practical implementation of MINRES can become quite complicated, we provide the algorithm
for the algebraically equivalent conjugate residual (CR) method \cite{Saad03} for brevity. The major computational steps (and costs) of
MINRES for our purposes can be easily highlighted in the CR algorithm.  In both Algorithms \ref{alg:cg} and \ref{alg:minres}, we let $tol$ denote the user-defined convergence
tolerance, $maxIter$ represent the max-allowable iterations, and $\bv^{(0)}$ be the initial guess.

\begin{algorithm}
\small
\caption{Conjugate Gradient Method for $H\bv = \bb$}\label{alg:cg}
\begin{algorithmic}[1]
\State $\textbf{r}^{(0)} = \bb-H\bv^{(0)}$
\State $\textbf{p}^{(0)} = \textbf{r}^{(0)}$
\State $i = 0$
\While{$i < maxIter$ and $\|\textbf{r}^{(i)}\|> tol$}
\State $\alpha_{i} = \frac{\|\textbf{r}^{(i)}\|^{2}}{(\textbf{p}^{(i)})^{T}H\textbf{p}^{(i)}} $
\State $\textbf{x}^{(i+1)} = \textbf{x}^{(i)} + \alpha_{i}\textbf{p}^{(i)}$
\State $\textbf{r}^{(i+1)} = \textbf{r}^{(i)} - \alpha_{i}H\textbf{p}^{(i)}$
\State $\beta_i = \frac{\|\textbf{r}^{(i+1)}\|^2}{\|\textbf{r}^{(i)}\|^2}$
\State $\textbf{p}^{(i+1)} = \textbf{r}^{(i+1)} + \beta_{i}\textbf{p}^{(i)}$
\State $i = i + 1$
\EndWhile
\end{algorithmic}
\end{algorithm}

\begin{algorithm}
\small
\caption{Conjugate Residual Method for $H\bv = \bb$}\label{alg:minres}
\begin{algorithmic}[1]
\State $\textbf{r}^{(0)} = \bb-H\bv^{(0)}$
\State $\textbf{p}^{(0)} = \textbf{r}^{(0)}$
\State $i = 0$
\While{$i < maxIter$ and $\|\textbf{r}^{(i)}\|> tol$}
\State $\alpha_{i} = \frac{(H\textbf{r}^{(i)})^T \textbf{r}^{(i)}}{\|H\textbf{p}^{(i)}\|^2} $
\State $\textbf{x}^{(i+1)} = \textbf{x}^{(i)} + \alpha_{i}\textbf{p}^{(i)}$
\State $\textbf{r}^{(i+1)} = \textbf{r}^{(i)} - \alpha_{i}H\textbf{p}^{(i)}$
\State $\beta_i = \frac{(H\textbf{r}^{(i+1)})^T\textbf{r}^{(i+1)}}{(H\textbf{r}^{(i)})^T\textbf{r}^{(i)}}$
\State $\textbf{p}^{(i+1)} = \textbf{r}^{(i+1)} + \beta_{i}\textbf{p}^{(i)}$
\State $i = i + 1$
\EndWhile
\end{algorithmic}
\end{algorithm}

In general, CG is used for SPD matrices, while MINRES is reserved for symmetric indefinite matrices.  While we provide a comparative analysis of these methods in the present study, the development of a specific recipe for when to choose one method over the other for
these applications is
part of ongoing work.  Our results demonstrate flexibility in the choice of solver,
depending on user preference, without sacrificing computational time.

We note that simple optimizations can be immediately taken advantage of in both algorithms.  Since both require the initial residual,
$\textbf{r}^{(0)}$, using an initial guess of $\textbf{v}^{(0)} = [0~0~\cdots 0]^T$ allows us to avoid
the matrix-vector product in line 1 of each algorithm and we simply let $\textbf{r}^{(0)} = \textbf{b}$. Further, and as we will demonstrate
in our experiments, basing termination of CG and MINRES on maximum iterations alone is
sufficient for generating an accurate Pareto front, allowing us to remove
the test for convergence in future implementations (i.e., the computation of $\|\textbf{r}^{(j)}\|$
at every iteration).  While this is not a significant cost alone, when solving a large number of linear
systems, the accumulated cost may no longer be negligible. Ongoing work focuses on continued improvements to our implementations to take advantage of these, and other, opportunities for speed-up in the convergence of these iterative methods.

We note that in our results (Section \ref{sec:experiments}), we provide data from experiments using CG with the Hessian matrix
{\it for comparison purposes only}. Because we cannot guarantee that the Hessian matrix is SPD, we would naturally choose MINRES as our
iterative solver.  As we later show, we still generate a similar Pareto front (in terms of quality) using CG with the Hessian, suggesting
that some of these matrices may be (close to) SPD or that CG does not suffer as dramatically as the theory suggests for
(slightly) indefinite matrices.  Though, we do highlight that, in many cases, CG with the Hessian results in the largest overall runtime. 

\subsection{Predictor-Corrector based on Gauss-Newton approximation}

Algorithms \ref{alg:PCGN} and \ref{alg:HVP} describe how we generate a Pareto front. After training an initial Pareto optimal
network with parameter $\bx$, we use Algorithm \ref{alg:PCGN} to explore the Pareto front in a breadth-first style. Algorithm \ref{alg:HVP} describes the Hessian-vector product computation using the Gauss-Newton approximation. Algorithm \ref{alg:HVP} is used at each iteration of MINRES and CG in the predictor
step of the Predictor-Corrector algorithm. We take advantage of the fact that we compute a weighted gradient of the objective
functions, so $J$ and $J^T$ can be represented as vectors. We can then use the Gauss-Newton approximation
to compute the product $H(\mathbf{x})\mathbf{v}$ using only a single gradient computation, followed by $m$ 
inner products and scalar-vector product. Note that in our method, we use $K = 1, \alpha = 0.1$ and predetermined values for $\beta$. However, one could also store and reuse the computed weighted gradient and instead randomly sample $\beta$ to generate more than 1 child networks from a single parent.

\begin{algorithm}
\small 
\caption{PC-GN}\label{alg:PCGN}
\begin{algorithmic}[1]
\State $\mathbf{x} = \text{An initial Pareto optimal network}$
\State $N = \text{Total number of networks to generate}$
\State $K = \text{Number of children to generate per network}$
\State $\beta$: moving directions of the objective function values.
\State $count = 0$
\State Initialize a queue $q$ and list $output$
\State Add $\mathbf{x}$ to $q$ and $output$
\While{$count < N$}
\State Pop a network $parent$ from $q$
\State $numChildren = 0$
\While{$numChildren < K$}
\State Use MINRES/CG to solve $H(\mathbf{x})\mathbf{v} = J^T\beta$.
\State $\mathbf{x} = \mathbf{x} + \alpha\mathbf{v}$
\State Correct $\mathbf{x}$ with one step of SMGD
\State optimize $child$ using a single training epoch
\State add $child$ to $q$ and $output$
\State $count = count + 1$
\State $numChildren = numChildren + 1$
\EndWhile
\EndWhile
\State repeat 8-18 for $\beta = (0, 1)^T$
\State remove dominated points from output
\end{algorithmic}
\end{algorithm}
\begin{algorithm}
\small 
\caption{HVP computation using GN}\label{alg:HVP}
\begin{algorithmic}[1]
\State $\mathbf{x} = \text{A Pareto optimal network}$
\State $\mathbf{v} = \text{The vector to multiply } H \text{ by}$
\State Use automatic differentiation to compute the gradients of the objective functions
\State $J = \text{weighted sum of gradients}$
\State $dot = \text{inner product } \langle J, \mathbf{v}\rangle $
\State \Return $H(\mathbf{x})\mathbf{v} = dot * J$
\end{algorithmic}
\end{algorithm}

\section{Experiments}
\label{sec:experiments}
\subsection{MOO tasks and datasets}
We use multiple MOO tasks and datasets to demonstrate the efficiency and effectiveness of the proposed GN-PC-MOO method.
Table~\ref{tab:tasks_datasets} shows the sizes of the datasets and the number of parameters in the corresponding neural networks.
\begin{table}
	\caption{Dataset and model sizes}
	\label{tab:tasks_datasets}
	\centering
	\begin{tabular}{r|ccccc}
		\toprule
		Datasets & $|\mathcal{V}^P|$ & $|\mathcal{V}^R|$ & $|\mathcal{V}^U|$  & Model size\\\hline
	 YelpChi & 201 & 67395 & 38063 & 1234 \\
		 YelpNYC & 923 & 358911 & 160220 & 1234 \\
		 YelpZip & 5044 & 608598 & 260277 & 1234 \\\hline
		\bottomrule
	\end{tabular}
\end{table}
\begin{figure*}
    \centering
    \includegraphics[width=1\textwidth]{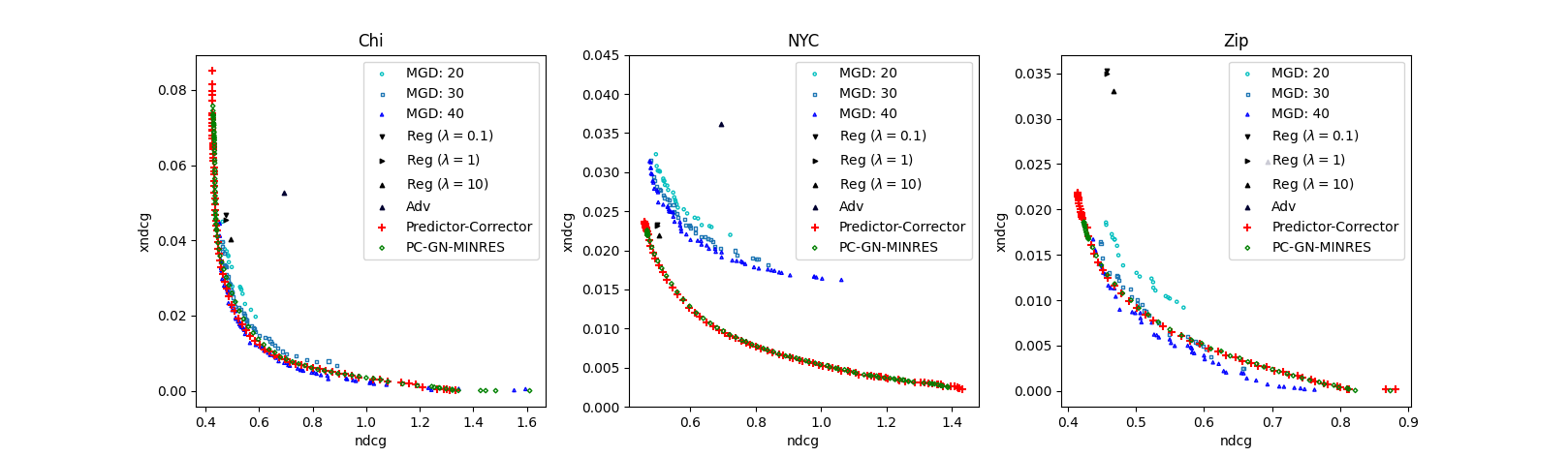}
    \caption{\small Pareto fronts found by various methods.
    The two axes represent $f_1$ and $f_2$ objective function values on the training set (same for Figure~\ref{fig:corrector},~\ref{fig:differ_solver},~\ref{fig:diff_iter_solvers_GN},\ref{fig:diff_iter_solvers_MINRES}, and~\ref{fig:corrector}.
    From left to right: comparison on YelpChi, YelpNYC, and YelpZip datasets.
    Within each figure: MGD is the SMGD method running for 20, 30, and 40 iterations;
    Reg ($\lambda$) means the fairness regularization method with a specific regularization parameter $\lambda$;
    Adv refers to the adversarial training method for fair machine learning;
    Predictor-Corrector uses the Hessian matrix with the CG solver,
    while the proposed method PC-GN-MINRES uses the Gauss-Newton approximation with the MINRES solver.
    }
    \label{fig:all_fronts}
\end{figure*}


\noindent\textbf{Fair fake review detection with GNN}.
Reviews on e-commerce, such as Amazon and Yelp, are important to customers and business owners, and there are many misleading fake reviews that need to be detected to ensure the trustworthiness of the reviews.
The review data can be represented as a review-graph defined as $\mathcal{G} = (\mathcal{V}, \mathcal{E})$, where $\mathcal{V}=\{v_1, \dots, v_N\}$ denotes the set of nodes and $\mathcal{E} \subseteq \mathcal{V} \times \mathcal{V}$ represents the set of undirected edges.
There are three types of nodes in $\mathcal{G}$, i.e., user, review, and product, respectively, and each node can be of only one of the three types. 
We denote the subsets of nodes of the three types as $\mathcal{V}^{U}, \mathcal{V}^{R}, \mathcal{V}^{P} \subset \mathcal{V}$, respectively.
Each node $v_i \in \mathcal{V}$ has a feature vector
$\mathbf{x}_i$, where the subscript is the node index.
The neighbor of the node $v_i$ is $\mathcal{N}(i) = \{v_j \in \mathcal{V}| e_{i, j}\in \mathcal{E}\}$.

GNN~\cite{kipf2016semi} is the state-of-the-art method for node prediction.
For each node, a GNN model summarizes a node's neighborhood via message passing to predict if a review node is fake (positive) or genuine (negative). 
We let $\hat{y}_i = h(v_i;\bx)$ 
be the predicted probability of node $v_i$ being fake,
and train the model by
minimizing the first objective, the cross-entropy loss
on the set of labeled training reviews $\mathcal{V}^{\text{Tr}}$ :
\begin{equation}
	\label{eq: cross-entropy loss}
	f_1(\bx) =- \frac{1}{l} \sum_{i=1}^l y_i \log \hat{y}_i + (1-y_i)\log(1-\hat{y}_i),
\end{equation}
where $y_i\in \{0, 1\}$ is the label of a labeled review node $v_i\in \mathcal{V}^{\text{Tr}}\cap \mathcal{V}^{\text{R}}$
and $l$ is the total number of such labeled review nodes.

We use the normalized discounted cumulative gain (NDCG) loss to measure the overall detection accuracy across all reviews:
\begin{equation}
\frac{1}{Z} \sum_{j=1}^{l} \mathds{1}[y_j=1]\frac{1}{\log(r_j^0+1)},
\end{equation}
where $r_i$ is the ranking position of the $i$-th labeled node
among all $m$ nodes and $Z$ is the maximal possible NDCG score as a normalization factor.
%

We also aim to reduce unfair treatment in the detection as the second objective.
In particular,
there are two groups of nodes, the favored group (indicated by $A=0$, consisting of reviewers each of which posts the most reviews)
and the remaining reviewers posting less reviews are in the protected group (indicated by $A=1$).
On the training data, the protected group are labeled as spammers more often,
biasing the trained GNN model 
to have a higher false positive rate over the protected group and leading to unfair detection.
To measure the discrepancy in the detection accuracies over the two groups,
we measure the NDCG on the two groups separately and take the absolute difference
between the two NDCG scores as the second objective function $f_2$.
Overall,
we aim to optimize the GNN model $h(\bx)$ to find Pareto fronts of classification loss \textit{vs.} detection discrepancy tradeoffs.





\subsection{Baselines and settings}
We have two regular fair machine learning baselines that can find a solution that aims to minimize both objectives.
Fairness regularization~\cite{Zafar2017,Muhammad2019,Kamishima2012} minimizes $f_1(\bx)+\lambda f_2(\bx)$ where the second term is for fairness regularization.
Adversarial training~\cite{Dai2021} trains an adversary that classify data into two groups, while our goal is to minimize $f_1$ while maximize the adversary's classification classification error.
We also include the following MOO methods as baselines.
\begin{itemize}[leftmargin=*]
	\item SMGD: the baseline proposed in~\cite{Burkholder2021}. It starts from several random initial solutions and then match towards a Pareto front, and it is supposed to be more time-consuming than the predictor-corrector methods as it can take many iterations to reach the front, though it use only first-order derivatives only.
	\item PC-Hessian-CG: use the Pearlmutter trick~\cite{Pearlmutter1994} to evaluate the Hessian-vector product and use CG to solve the linear system for finding an exploration step in the predictor.
	If the Hessian matrices are PSD, the CG should be a suitable choice.
	Though there is no guarantee that the Hessian matrices will be PSD, we include this baseline for a comprehensive comparison.
			The corrector use multiple SMGD steps.
	\item PC-Hessian-MINRES: same as PC-Hessian-CG, except that the CG method is replaced with a MINRES solver.
\end{itemize}
We compare these baselines to two variants of the proposed method, PC-GN-CG and PC-GN-MINRES, that use Gauss-Newton approximation of the Hessian matrices. 
It is expected that the Gauss-Newton approximation will reduce the running time of methods that rely on Hessian-vector products, as we analyzed in Section~\ref{sec:gn}.
while we use different solvers to study whether the properties of the linear systems can influence the number of iterations, the running time, and the Pareto front quality.

\noindent\textbf{Hyperparameter setting}.
We vary the $\lambda$ in $f_1(\bx)+\lambda f_2(\bx)$ for the fairness regularization method.
For the PC methods,
we generate one initial solution using 75 optimization steps by SMGD,
followed by 100 predictor-corrector steps in both directions ($\beta=[-1,1]^\top$ and $\beta=[1,-1]^\top$).
For the predictor we used a step size of 0.1 and 50 max iterations for the solvers following the ablation studies explored in \cite{ma20aICML},
and for the corrector we used a step size of 0.01.
We ran SMGD~\cite{Liu2019} for different number of expanding iterations (20, 30, and 40 epochs).
The best step size was found to be 0.005 for the descent step.

\subsection{Pareto front Quality}
The closer a Pareto front to the minimal values of individual objectives, the better.
From Figure~\ref{fig:all_fronts}, 
We have the following observations.
\begin{itemize}[leftmargin=*]
	\item
As shown in Figure~\ref{fig:all_fronts}, 
on the YelpChi dataset, 
the two PC methods (Predictor-Corrector and PC-GN-MINRES) are comparable in the quality of fronts produced by SMGD running for 40 epochs (which takes much longer time as we show in Section~\ref{sec:running_time}).
On the YelpNYC dataset, the front produced by the predictor-corrector methods completely dominates the fronts produced by the SMGD methods even with 40 epoches.
On the YelpZip dataset, the fronts produced by the PC methods is slightly dominated by SMGD that runs for 40 epochs, which however has a much longer running time.
	\item
		The spread of the fronts generated by SMGD with 20 epochs is much smaller than those from the predictor-corrector methods, which spend similar amount of time as the SMGD method. 
		Overall, our method is a more efficient Pareto front generator and is able to find quality solutions at a faster rate than the multi-gradient descent algorithm.
		\item Both the regularization and adversarial training methods are domininated by SMGD and the PC methods. Furthermore, each baseline can generate just one solution and thus does not offer users to pick a desired final solution. 
\end{itemize}

\begin{figure*}
    \centering
    \begin{subfigure}[b]{0.21\textwidth}
        \centering
        \includegraphics[width=\textwidth]{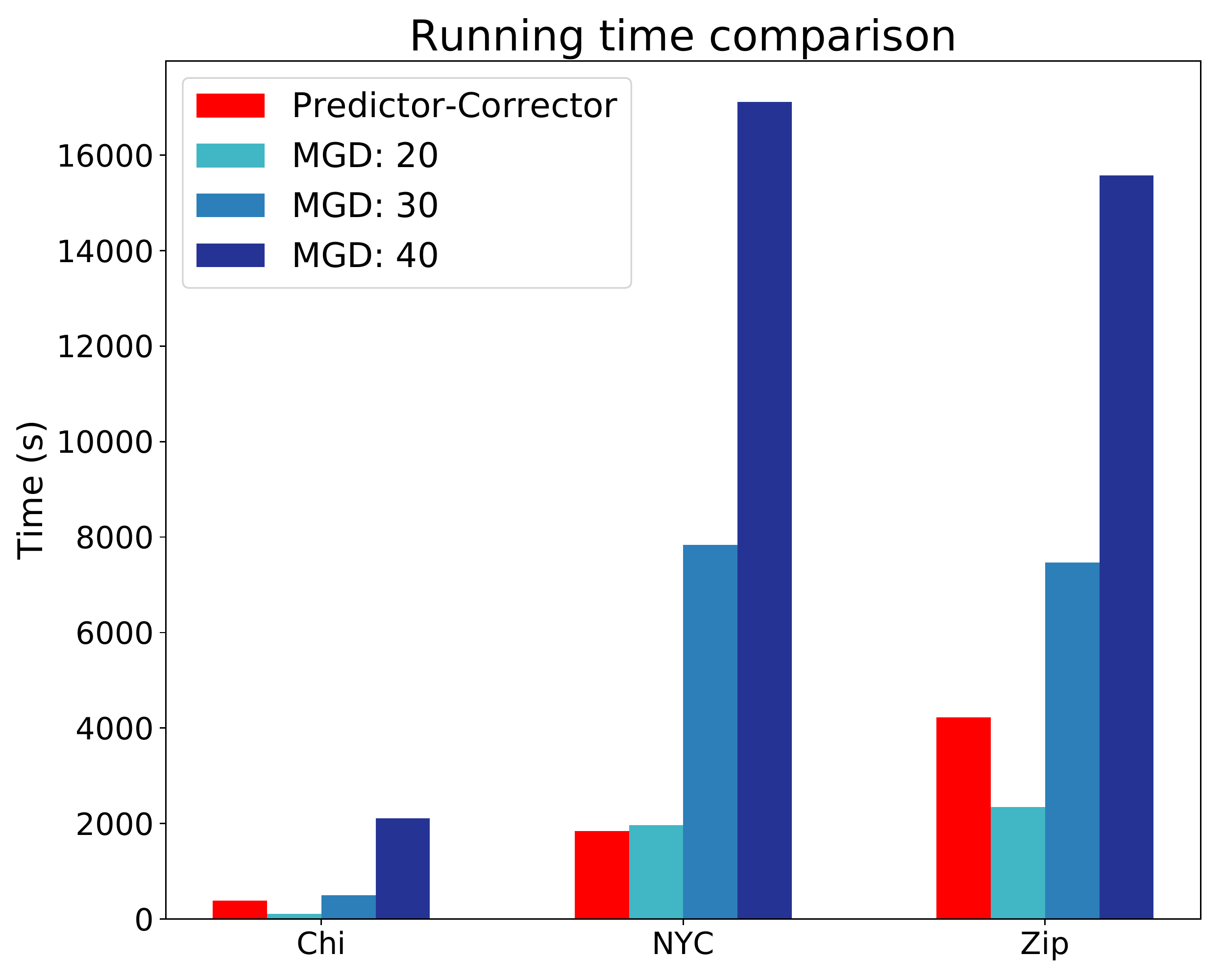}
         \caption{Running time of PC-Hessian-CG and SMGD}
         \label{fig:running_time_SMGD_PC}
     \end{subfigure}\hfill
    \begin{subfigure}[b]{0.25\textwidth}
     \centering
        \includegraphics[width=\textwidth]{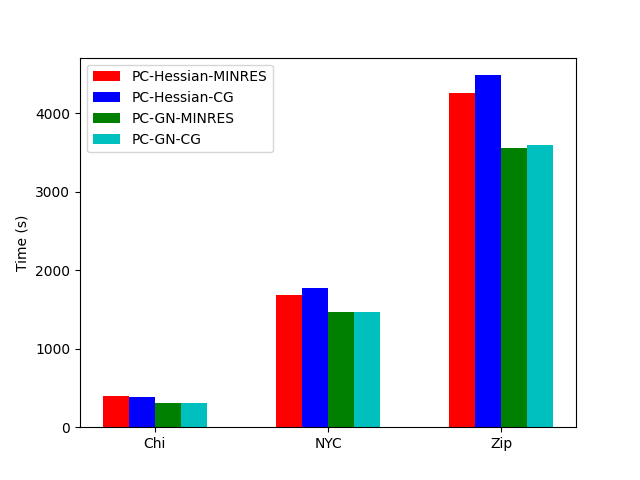}
         \caption{Running time of PC methods with 10 iterations for the solvers.}
         \label{fig:running_time_PC_10_iters}
    \end{subfigure}\hfill
    \begin{subfigure}[b]{0.25\textwidth}
     \centering
        \includegraphics[width=\textwidth]{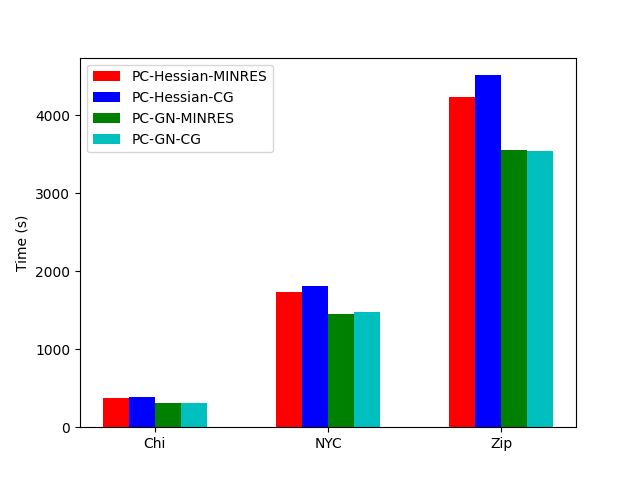}
         \caption{Running time of PC methods with 25 iterations for the solvers.}
         \label{fig:running_time_PC_25_iters}
    \end{subfigure}\hfill
    \begin{subfigure}[b]{0.25\textwidth}
     \centering
        \includegraphics[width=\textwidth]{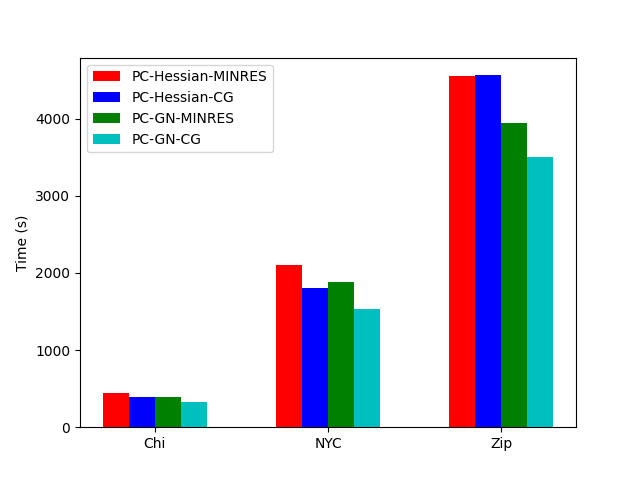}
         \caption{Running time of PC methods with 50 iterations for the solvers.}
         \label{fig:running_time_PC_50_iters}
    \end{subfigure}
    \caption{\small \textit{Left}: Running time of Predictor-Corrector (PC) and SMGD on the three datasets for fake review detection.
    From (b)-(d): comparing the running time of Predictor-Corrector (PC) methods with Hessian and Gauss-Newton approximation and CG/MINRES solvers, with 10, 25, and 50 iterations for the solver to solve the linear system in Eq. (\ref{eq:predictor_direction}).}
    \label{fig:running_time}
\end{figure*}

\subsection{Algorithm Speed}
\label{sec:running_time}
We demonstrate how the proposed Gauss-Newton approximation reduces the running time of the PC methods.
	We control the quality of the generated fronts and measure the time needed for each methods to reach the fronts within an approximity.
		Based on Figure~\ref{fig:all_fronts},
		we regard the fronts found by SMGD as the target front and measure how long it takes for other methods to reach those fronts.
		As shown in Figure~\ref{fig:running_time_SMGD_PC},
			the method PC-Hessian-CG has substantial speed ups over the 
			multi-gradient descent method with over a 5-fold of speedup on YelpChi,
			a 9-fold of speedup on YelpNYC,
			and over a 3-fold speedup on YelpZip.
		We further compare the running time of PC-GN-CG and PC-GN-MINRES that use Gauss-Newton approximation against PC-Hessian-CG and PC-Hessian-MINRES that use the Hessian matrices.
		From Figures~\ref{fig:running_time_PC_10_iters}-\ref{fig:running_time_PC_50_iters} with different number of iterations for the solvers (CG or MINRES) to solve the linear system Eq. (\ref{eq:predictor_direction}),
		we can see the the running time is further reduced from that of PC-Hessian-CG.
		This indicates that the Gauss-Newton approximation can reach the same Pareto fronts significantly faster.
		Note that the CG solver most of the time slightly runs faster than the MINRES solver when Gauss-Newton approximation is used,
		but can be slower than MINRES when the Hessian is used. We again reiterate that since we cannot guarantee the Hessian is PSD, we would employ MINRES in practice and provide these results for comparison purposes only.
		Therefore, that CG runs slower in this case is not unexpected.

\begin{figure*}
    \centering
    \includegraphics[width=0.8\textwidth]{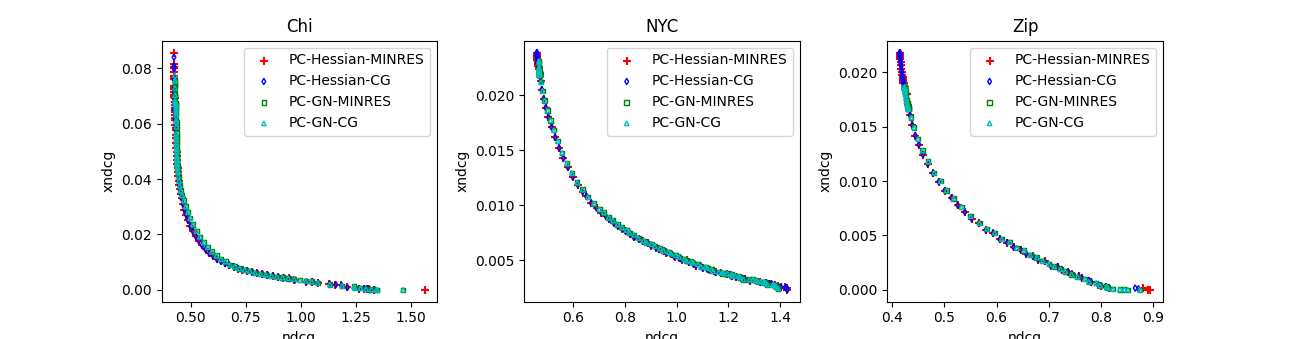}
    \caption{\small Comparison of the quality of the Pareto front when using Hessian and Gauss-Newton matrices with different solvers.}
    \label{fig:differ_solver}
\end{figure*}

\begin{figure*}
\centering
   \begin{subfigure}[b]{0.8\textwidth}
         \centering
         \includegraphics[width=\textwidth]{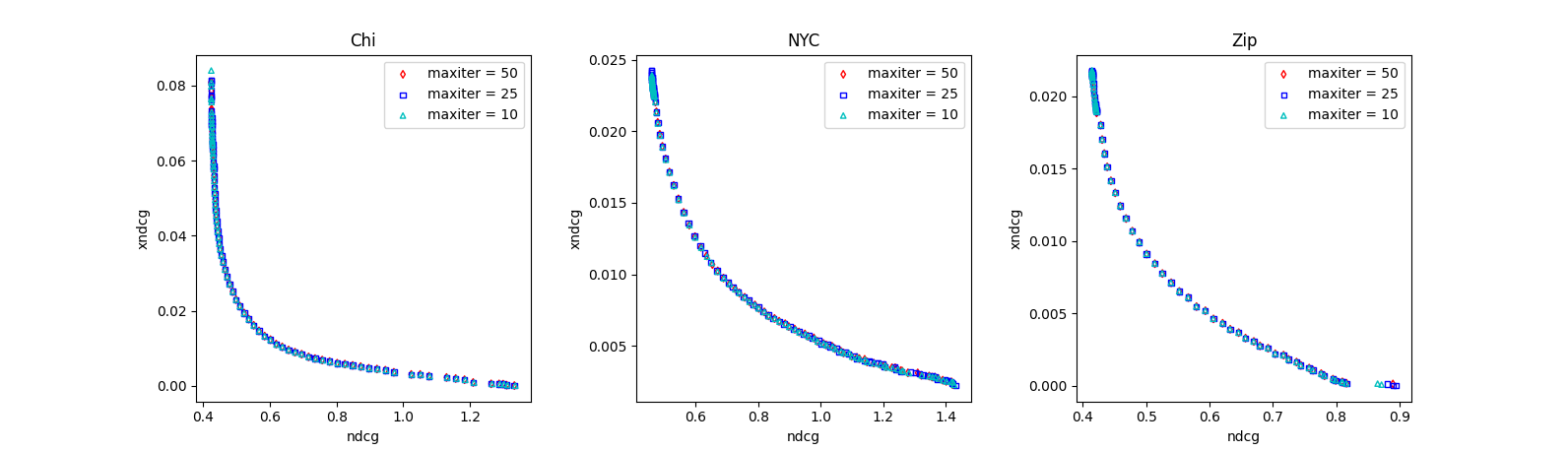}
         \caption{PC-Hessian-CG}
         \label{fig:diff_iter_solver_PC_Hessian_CG}
     \end{subfigure}
     
    \begin{subfigure}[b]{0.8\textwidth}
        \centering
        \includegraphics[width=\textwidth]{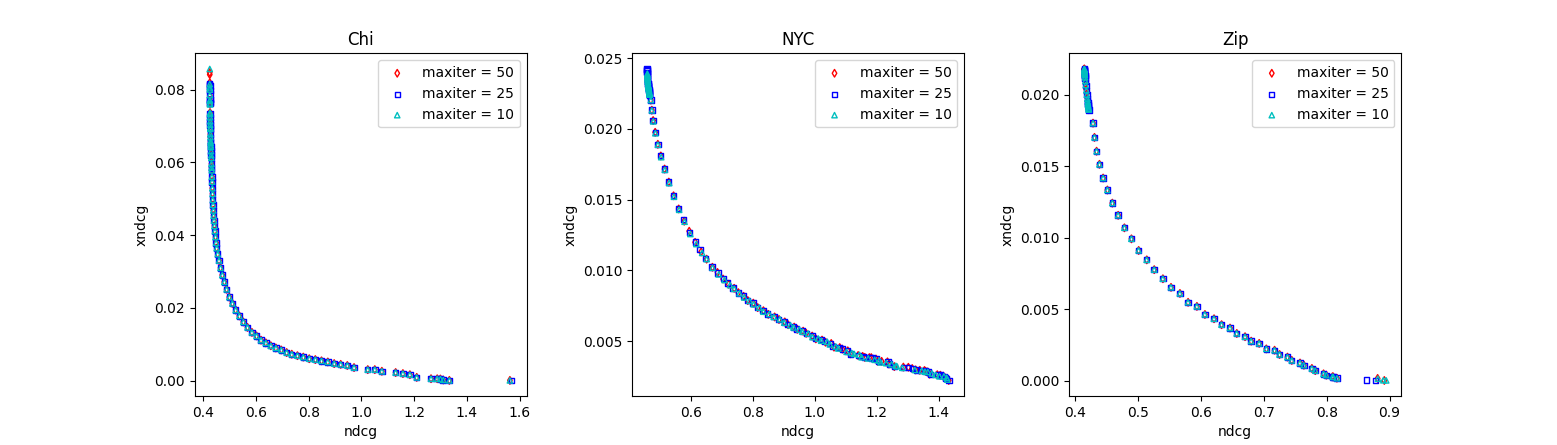}
        \caption{PC-Hessian-MINRES}
        \label{fig:diff_iter_solver_PC_Hessian_MINRES}
    \end{subfigure}
      \caption{\small Pareto front quality when varying the number of maximum iterations set for the linear solvers with Hessian matrices.}
  \label{fig:diff_iter_solvers_MINRES}
\end{figure*}

\begin{figure*}
\centering
   \begin{subfigure}[b]{0.8\textwidth}
  \includegraphics[width=\textwidth]{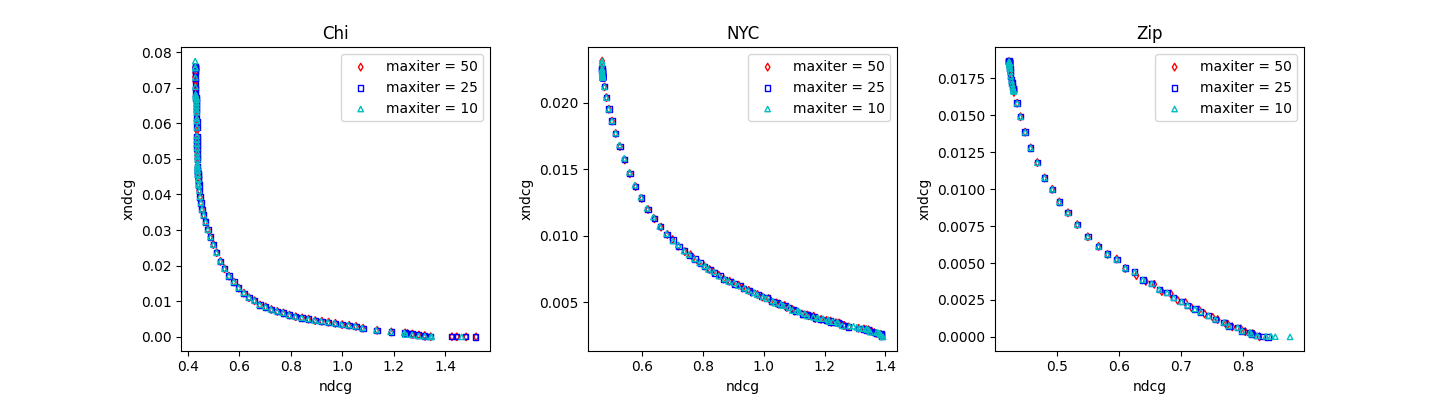}
     \caption{PC-GN-CG}
     \label{fig:diff_iter_solver_PC_GN_CG}
 \end{subfigure}
 
\begin{subfigure}[b]{0.8\textwidth}
  \includegraphics[width=\textwidth]{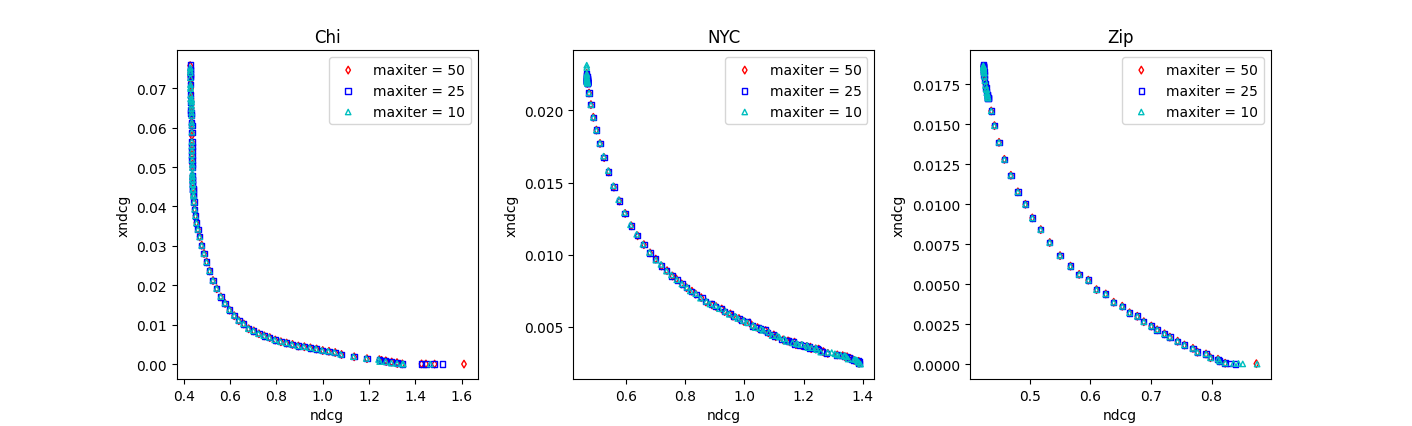}
    \caption{PC-GN-MINRES}
    \label{fig:diff_iter_solver_PC_GN_MINRES}
\end{subfigure} 
  \caption{\small Pareto front quality when varying the number of maximum iterations set for the linear solvers with Gauss-Newton approximations.}
  \label{fig:diff_iter_solvers_GN}
\end{figure*}


\subsection{Ablation and sensitivity studies}
\begin{figure*}
  \centering
  \includegraphics[width=0.3\textwidth]{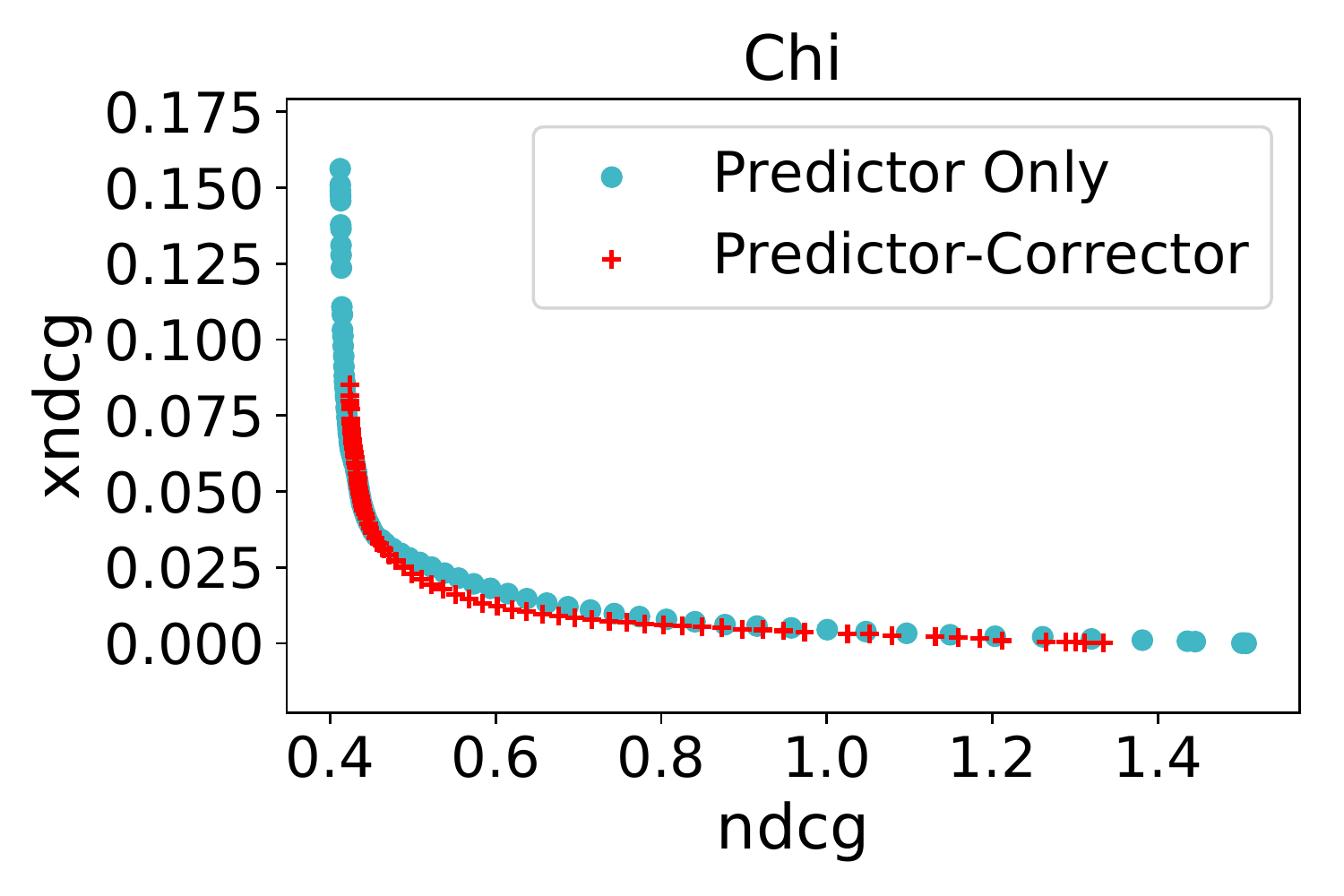}
  \includegraphics[width=0.3\textwidth]{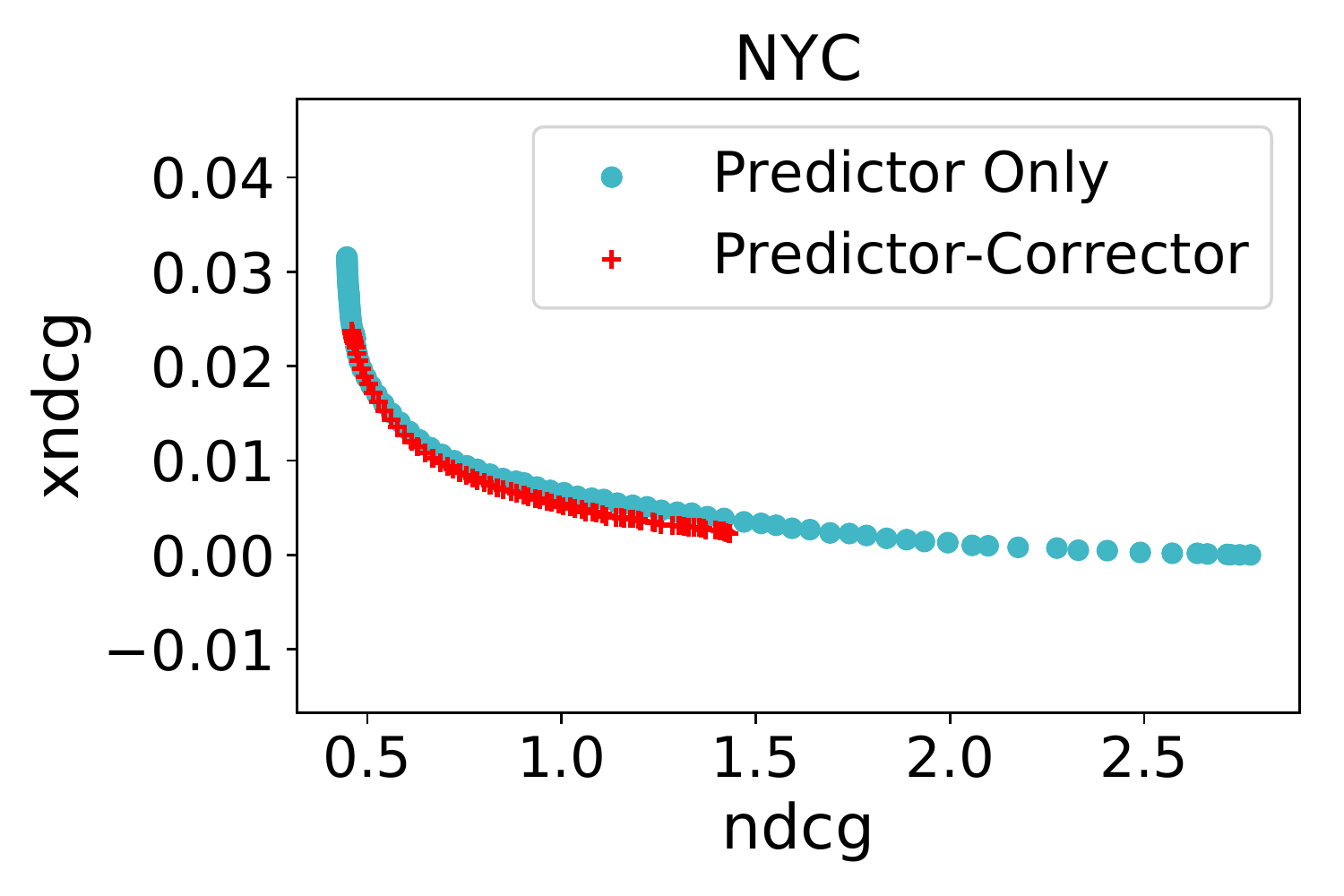}
  \includegraphics[width=0.3\textwidth]{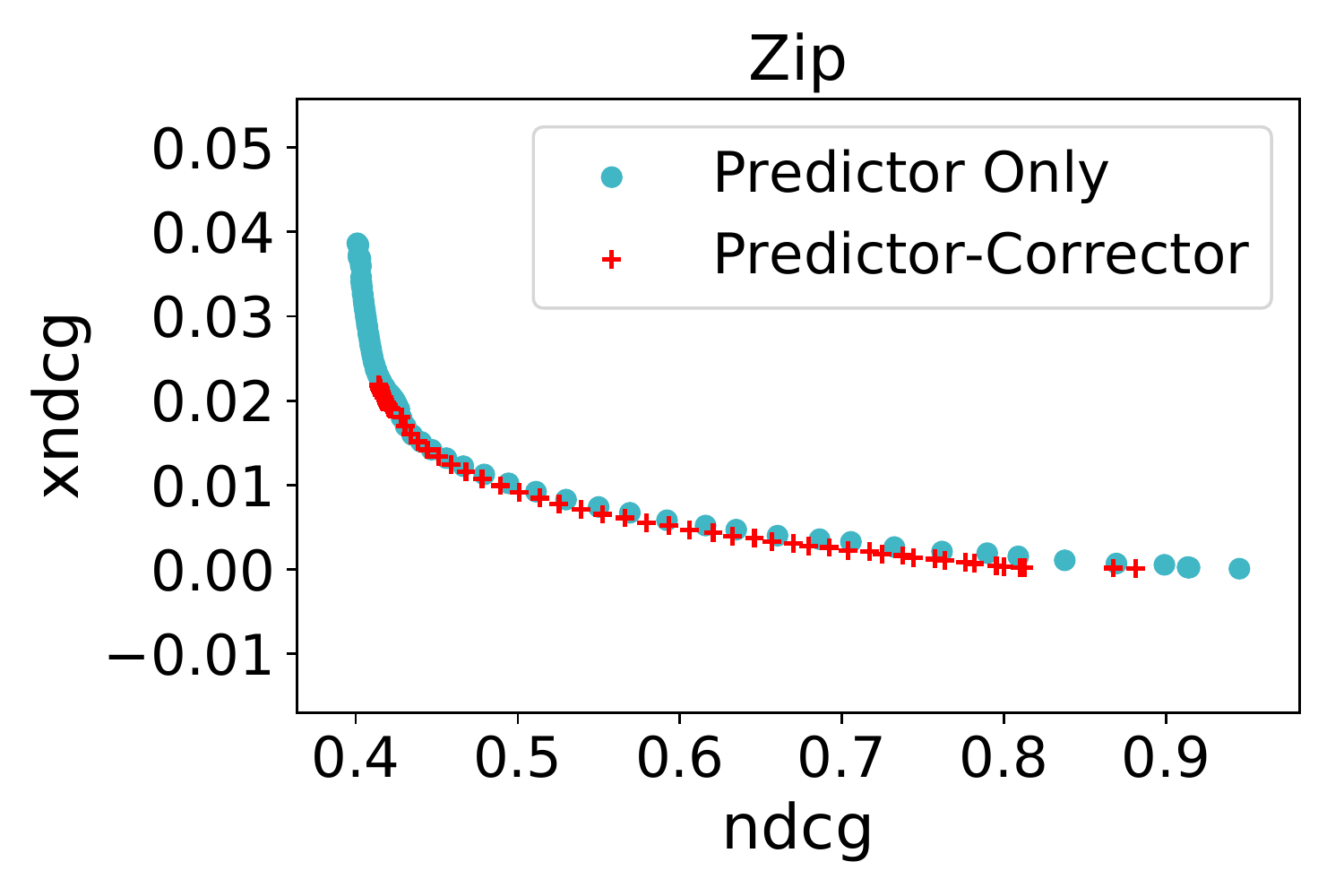}
  \caption{\small Ablation study performed to analyze the impact of the corrector step.
}
  \label{fig:corrector}
\end{figure*}

There are several components and hyperparameters of the proposed methods,
		and we will study how they influence the performance (quality of the Pareto fronts and running time).
\begin{itemize}[leftmargin=*]
	    
	\item \textbf{Comparing different solvers.}
	In Figure~\ref{fig:running_time}, we show the total running time when using PC-Hessian-MINRES, PC-Hessian-CG, PC-GN-MINRES, and PC-GN-CG for varying maximum iterations, and in
	Figure~\ref{fig:differ_solver} we show the quality of the Pareto fronts with maximum iterations set to 10.  We observe that, in general, the Gauss-Newton approximation results in an overall faster runtime, but that
	there is no significant difference when using CG or MINRES.  We also see that the running time does not change meaningfully when we reduce the number of maximum iterations for CG and MINRES 
	(for all tests).  
	    This indicates that (1) the Gauss-Newton approximation is quite accurate while significantly reducing running time of the solvers, and (2) that the user can safely choose either CG or MINRES depending on preference (or availability) without affecting these same metrics.
	
	\item \textbf{The quality of the Pareto fronts with different maximum iterations for the solvers.}
	    We compare the quality of the Pareto fronts found by our methods with a varied number of steps of the solvers that find the exploration direction $\bv$ as in Eq. (\ref{eq:predictor_direction}).
	    Each row of figures in
	    Figures~\ref{fig:diff_iter_solvers_MINRES} and ~\ref{fig:diff_iter_solvers_GN} show that for a fixed combination of linear system and solver, for a varying number of the maximum iterations for both CG and MINRES on three datasets,
	    we generate more or less the same Pareto fronts.
	     This indicates that we can set a very modest number of maximum iterations for the iterative solver without affecting the quality of the Pareto fronts.
\item \textbf{The effect of the corrector step.}
	    We investigate whether the corrector is necessary after each predictor step,
		or if the predictor's approximation of the tangent direction is good enough and the corrector is unnecessary.
		We run PC-Hessian-CG with and without the corrector.
		The results can be seen in Figure \ref{fig:corrector}.
		Without the corrector step, the algorithm produces fronts that are slightly dominated by those generated with a corrector.
		However, the corrector also seems to shrink the Pareto fronts (shown in red) towards the bottom left corner, while more Pareto optima can be produced extending the fronts produced by the PC-Hessian-CG, indicating that the predictor can indeed explore the manifold of Pareto optima to find new optimal solutions.
		In summary, the addition of a corrector step seems to trade off some spread of the produced solutions for more reduction in the objective function values where the objectives have the most competition.
		Therefore, we choose to utilize the corrector in all of the experiments.
\end{itemize}

\section{Related Work}
Multi-objective optimization has been studied extensively in numerical optimization~\cite{Fliege2000,MERCIER2018808,Kais99,Fliege2018,Martin2018,Liu2019} and been applied to machine learning problems, such as multi-task learning and fair machine learning~\cite{Sener2018nips,Lin2019_nips,mahapatra20a,ma20aICML}.
We focus on the predictor-corrector method for its efficiency.
Different from ~\cite{ma20aICML,Martin2018} that directly use CG and Pearlmutter trick to solve $H\mathbf{v}=\bb$,
we improve its efficiency by exploiting low-rank approximation of the Hessian matrix $H$ and various more advanced iterative methods such as MINRES.

Solving a linear system for a descent direction $\textbf{v}$ to optimize a machine learning model is commonly found in second-order numerical optimization.
Second-order methods can exploit the local curvature information of an objective function, $\textbf{f}$, to properly scale the gradient vector for a more robust descent direction,
while first-order methods need to fine-tune the step size in the direction of the negative gradient and can be time-consuming in practice.
The Newton method is a second-order method that solves the linear system $H\textbf{v}=\nabla \textbf{f}$, using iterative methods such as conjugate gradient.
With many parameters to update, as it is typical for neural networks, the Hessian matrix cannot be computed and stored explicitly, and the Pearlmutter trick is proposed to rely on backpropagation for fast Hessian-vector product evaluation to solve for $p$.
Another issue with using $H$ is that it is not guaranteed to be positive definite and the CG method is not guaranteed to converge to the optimal descent direction.
Gauss-Newton and Fisher information matrices are low-rank SPD matrices approximating the Hessian $H$.
These approximating matrices are not only PSD, but also make the matrix-vector with $\textbf{v}$ much cheaper to compute without calling backpropagation for each iteration of CG that evaluate $H\textbf{v}$.
Rather, the approximating matrices are in the form of a sum of outer products of vectors that can be computed and stored explicitly.
Besides solving $H\textbf{v}=\nabla \textbf{f}$ iteratively, in the work~\cite{Martens2015ICML},
block diagonal matrices approximation of $H$ are derived so that direct inverses of the block diagonal matrices are less expensive to compute.
All the previous work optimizes a single objective and did not aim to speed up the predictor-corrector framework for MOO.


Other empirical and theoretical comparative analyses of MINRES and CG have been performed (see e.g., \cite{Fong2012}). Previous work has also demonstrated the effectiveness of MINRES for MOO (see e.g., \cite{ma20aICML}) as it is a matrix-free solver that guarantees a monotonic decrease in the norm of the residual (of the approximation solution for (\ref{eq:predictor_direction})).  However, to our knowledge,
the present study will be the first analysis of the computational time and quality of the Pareto front when using CG and MINRES for applications arising in multiple-objective optimization.  In \cite{ma20aICML}, a comparative analysis is performed when varying the maximum number of iterations of the iterative solver. However, in the present study, we show that we can achieve an accurate Pareto front with even fewer maximum iterations than those considered in \cite{ma20aICML}.

\section{Conclusion}
We aim to find Pareto fronts that represent multiple trade-offs between multiple objective functions.
Previous MOO methods start from a random initial solution or rely on second-order derivatives and are thus inefficient.
We propose a first-order approach called Gauss-Newton approximation to remove the need of the second-order derivatives, and embed the approach in the Predictor-Corrector method that can generate Pareto fronts by exploring the manifold with high efficiency.
We applied the method to a fake review detection task on three datasets, and demonstrated the proposed methods can find high-quality Pareto fronts using less time.
\bibliographystyle{plain}
\bibliography{references}

\begin{thebibliography}{10}

\bibitem{Burkholder2021}
Kai Burkholder, Kenny Kwock, Jiaxin Liu, and Sihong Xie.
\newblock Certification and trade-off of multiple fairness criteria in
  graph-based spam detection.
\newblock In {\em CIKM 2021}, 2021.

\bibitem{Dai2021}
Enyan Dai and Suhang Wang.
\newblock {Say No to the Discrimination: Learning Fair Graph Neural Networks
  with Limited Sensitive Attribute Information}.
\newblock In {\em WSDM}, 2021.

\bibitem{Fliege2018}
J.~Fliege, A.I.F Vaz, and L~N Vicente.
\newblock {Complexity of gradient descent for multiobjective optimization}.
\newblock Technical report, 2018.

\bibitem{Fliege2000}
J{\"{o}}rg Fliege and Benar~Fux Svaiter.
\newblock {Steepest descent methods for multicriteria optimization}.
\newblock {\em Mathematical Methods of Operations Research}, 51(3):479--494,
  2000.

\bibitem{Fong2012}
David Chin-Lung Fong and Michael Saunders.
\newblock Cg versus minres: An empirical comparison.
\newblock {\em SQU Journal for Science}, 2012.

\bibitem{Gar2020}
M.~Gargiani, A.~Zanelli, M.~Diehl, and F.~Hutter.
\newblock On the promise of the stochastic generalized {Gauss-Newton} method
  for training {DNN}s.
\newblock 2020.

\bibitem{Hestenes1952}
Magnus~R. Hestenes and Eduard Stiefel.
\newblock Methods of conjugate gradients for solving linear systems.
\newblock {\em Journal of research of the National Bureau of Standards},
  49:409--435, 1952.

\bibitem{Kais99}
Miettinen Kaisa.
\newblock {\em {Nonlinear Multiobjective Optimization}}, volume~12 of {\em
  International Series in Operations Research \& Management Science}.
\newblock Kluwer Academic Publishers, Boston, USA, 1999.

\bibitem{Kamishima2012}
Toshihiro Kamishima, Shotaro Akaho, Hideki Asoh, and Jun Sakuma.
\newblock {Fairness-Aware Classifier with Prejudice Remover Regularizer}.
\newblock In {\em Machine Learning and Knowledge Discovery in Databases}, pages
  35--50, 2012.

\bibitem{kipf2016semi}
Thomas~N Kipf and Max Welling.
\newblock Semi-supervised classification with graph convolutional networks.
\newblock {\em arXiv preprint arXiv:1609.02907}, 2016.

\bibitem{Lin2019_nips}
Xi~Lin, Hui-Ling Zhen, Zhenhua Li, Qing-Fu Zhang, and Sam Kwong.
\newblock {Pareto Multi-Task Learning}.
\newblock In {\em NeurIPS}, volume~32, 2019.

\bibitem{Liu2019}
Suyun Liu and L~N Vicente.
\newblock {The stochastic multi-gradient algorithm for multi-objective
  optimization and its application to supervised machine learning}.
\newblock Technical report, 2019.

\bibitem{ma20aICML}
Pingchuan Ma, Tao Du, and Wojciech Matusik.
\newblock {Efficient Continuous Pareto Exploration in Multi-Task Learning}.
\newblock In {\em ICML}, 2020.

\bibitem{mahapatra20a}
Debabrata Mahapatra and Vaibhav Rajan.
\newblock {Multi-Task Learning with User Preferences: Gradient Descent with
  Controlled Ascent in Pareto Optimization}.
\newblock In {\em ICML}, 2020.

\bibitem{Martens2015ICML}
James Martens and Roger Grosse.
\newblock {Optimizing Neural Networks with Kronecker-Factored Approximate
  Curvature}.
\newblock ICML, 2015.

\bibitem{Martin2018}
Adanay Mart{\'{i}}n and Oliver Sch{\"{u}}tze.
\newblock {Pareto Tracer: a predictor–corrector method for multi-objective
  optimization problems}.
\newblock {\em Engineering Optimization}, 2018.

\bibitem{MERCIER2018808}
Quentin Mercier, Fabrice Poirion, and Jean-Antoine D{\'{e}}sid{\'{e}}ri.
\newblock {A stochastic multiple gradient descent algorithm}.
\newblock {\em European Journal of Operational Research}, 271(3):808--817,
  2018.

\bibitem{Noc2006}
Jorge Nocedal and Stephen~J Wright.
\newblock {\em Numerical optimization}.
\newblock Springer, 2006.

\bibitem{Pearlmutter1994}
Barak~A Pearlmutter.
\newblock {Fast Exact Multiplication by the Hessian}.
\newblock {\em Neural Comput.}, 6(1):147--160, 1994.

\bibitem{Saad03}
Yousef Saad.
\newblock {\em Iterative Methods for Sparse Linear Systems, 2nd Ed.}
\newblock SIAM, 2003.

\bibitem{Saad86}
Yousef Saad and Martin~H. Schultz.
\newblock {GMRES}: a generalized minimal residual algorithm for solving
  nonsymmetric linear systems.
\newblock {\em SIAM Journal on Scientific and Statistical Computing},
  7(3):856--869, 1986.

\bibitem{sagun2017empirical}
Levent Sagun, Utku Evci, V~Ugur Guney, Yann Dauphin, and Leon Bottou.
\newblock Empirical analysis of the hessian of over-parametrized neural
  networks.
\newblock {\em arXiv preprint arXiv:1706.04454}, 2017.

\bibitem{Sener2018nips}
Ozan Sener and Vladlen Koltun.
\newblock {Multi-Task Learning as Multi-Objective Optimization}.
\newblock In {\em Advances in Neural Information Processing Systems},
  volume~31, 2018.

\bibitem{VanDerVorst2003}
Henk~A. van~der Vorst.
\newblock {\em GMRES and MINRES}, page 65–94.
\newblock Cambridge Monographs on Applied and Computational Mathematics.
  Cambridge University Press, 2003.

\bibitem{Wu2019}
Yongkai Wu, Lu~Zhang, and Xintao Wu.
\newblock {On Convexity and Bounds of Fairness-Aware Classification}.
\newblock In {\em The World Wide Web Conference}, WWW '19, 2019.

\bibitem{Zafar2017}
Muhammad~Bilal Zafar, Isabel Valera, Manuel {Gomez Rodriguez}, and Krishna~P
  Gummadi.
\newblock {Fairness Beyond Disparate Treatment and Disparate Impact: Learning
  Classification Without Disparate Mistreatment}.
\newblock In {\em WWW}, 2017.

\bibitem{Muhammad2019}
Muhammad~Bilal Zafar, Isabel Valera, Manuel Gomez-Rodriguez, and Krishna~P.
  Gummadi.
\newblock Fairness constraints: A flexible approach for fair classification.
\newblock {\em Journal of Machine Learning Research}, 2019.

\end{thebibliography}
\end{document}